# Design and Structural Validation of a Micro-UAV with On-Board Dynamic Route Planning

Inbazhagan Ravikumar[1†][0009-0003-4297-2289], Ram Sundhar[1†][0009-0002-3648-4716], Narendhiran Vijayakumar[1†*][0009-0004-1852-3368]

[1]National Institute of Technology, Tiruchirappalli, 620015, India
{inbazhagan.nitt,ramsundharnitt,narendhiranv.nitt}@gmail.com

**Abstract.** Micro aerial vehicles are becoming increasingly important in search and rescue operations due to their agility, speed, and ability to access confined spaces or hazardous areas. However, designing lightweight aerial systems presents significant structural, aerodynamic, and computational challenges. This work addresses two key limitations in many low-cost aerial systems under two kilograms: their lack of structural durability during flight through rough terrains and inability to replan paths dynamically when new victims or obstacles are detected. We present a fully customised drone built from scratch using only commonly available components and materials, emphasising modularity, low cost, and ease of assembly. The structural frame is reinforced with lightweight yet durable materials to withstand impact, while the onboard control system is powered entirely by free, open-source software solutions. The proposed system demonstrates real-time perception and adaptive navigation capabilities without relying on expensive hardware accelerators by offering an affordable and practical solution for real-world search and rescue missions.

## 1 Introduction

Micro-class quadcopters (≤2 kg) are often used in search-and-rescue (SAR) missions due to their agility and rapid deployment [1]. However, they face structural and navigational challenges, especially in rugged terrain. Lightweight designs often compromise stiffness and impact resistance, increasing failure risks during hard landings.

A structurally efficient quadcopter frame was developed, combining lightweight materials with improved impact resistance and aerodynamic shaping. The design was validated through structural and aerodynamic analyses to ensure reliability in demanding conditions [2].

Object detection models for different UAV applications predominantly involve trade-offs between one-stage (e.g., YOLO variants) and two-stage (e.g., Faster R-CNN) detectors. Lightweight CNNs such as SlimYOLOv3 enable efficient real-time victim detection on embedded platforms [3]. Combined with dynamic path replanning algorithms, ranging from greedy nearest-neighbour heuristics to optimal travelling-salesman solvers, drones can swiftly update their routes based on live detections, markedly enhancing mission responsiveness [4].

---

[†]These authors contributed equally to this work.
*Corresponding author: narendhiranv.nitt@gmail.com

This paper proposes an integrated quadcopter solution combining robust structural design with optimal onboard vision-driven autonomy, validated through structural analyses and experimental SAR scenario testing.

## 2    Related Work

This section explores prior research and technologies relevant to developing autonomous search and rescue (SAR) drones by examining existing work in three key areas: structural resilience and rigidity in UAV design, onboard vision systems and component selection, and dynamic path planning algorithms.

### 2.1    Structural Resilience and Rigidity

Drone frame design now emphasises structural resilience using lightweight composites like carbon fibre for their high strength-to-weight ratio [5]. Features like flexible landing gear and hybrid polymer joints improve shock absorption during rough landings. Finite Element Analysis (FEA) is widely used to evaluate stress and vibration behaviour, ensuring reliability across various conditions. However, balancing rigidity, weight reduction, and ease of manufacturing remains a key design challenge [6].

### 2.2    Onboard Vision and System Selection

Computer vision models for UAVs predominantly involve trade-offs between one-stage (e.g., YOLO) and two-stage (e.g., Faster R-CNN) detectors. Two-stage algorithms generally deliver higher accuracy but impose prohibitive computational loads on embedded UAV hardware [7]. By contrast, one-stage detectors achieve a more suitable speed–accuracy balance for real-time flight. Benchmark studies on platforms such as NVIDIA Jetson and Raspberry Pi series reveal marked differences in inference throughput and power efficiency, guiding optimal subsystem selection for onboard deployment [8]. Lightweight communication protocols like MAVLink efficiently handle telemetry and command exchange between UAV and ground station under strict energy constraints [9].

### 2.3    Dynamic Path Planning

Dynamic path planners for Search and Rescue UAVs range from ultra-light greedy schemes to heavy optimisation. Nearest-neighbour routes refined only by minor post-processing enable millisecond replans but inflate tour length by 15 % as target density rises [10]. MILP-based (Mixed-Integer Linear Programming) coverage formulations like the CUAVRP algorithm [11] minimise total distance yet become intractable once waypoint counts climb past a few dozen, breaching real-time limits. Meta-heuristic multi-objective strategies cut cost but still demand seconds to minutes of CPU time on embedded hardware [12]. This raises the concern about balancing the need for computational efficiency in path planning algorithms.

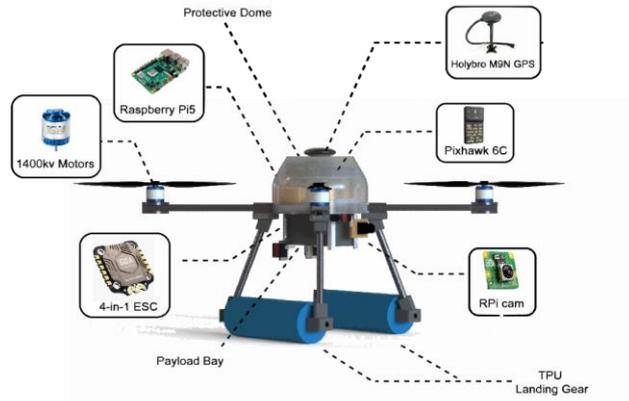

**Fig. 1** Visual abstract of the custom SAR quadcopter and its core hardware components.

## 3     Methodology

This section details the methodology for designing and developing autonomous search and rescue (SAR) drones. We cover structural design, onboard vision and system selection, and dynamic path planning.

### 3.1    Structural Resilience and Rigidity

The main objective of the structural design was to create a lightweight, rigid, and durable frame capable of withstanding flight loads, landing impacts, and collisions.

A quadrotor with an X-frame configuration was chosen for stability and sleeker design. The structure used a cross-frame layout with a double-layer sandwich hub, allowing compact component placement and high bending stiffness. Four-leg landing gear was designed to withstand drops from height and enable landing on uneven and slant terrain, as per the operation to be carried out.

Carbon fibre composites were used for the hub and extrusion arms due to their high strength-to-weight ratio. TPU was selected for 3D-printed landing gear shoes with gyroid infill pattern for impact absorption. A vacuum-formed PETG dome protects the electronics and reduces drag. Various structural analyses were carried out to optimise the weight and strength of each mechanical component, and computational fluid dynamics was carried out to study fluid-body interactions.

The airframe was tested under simulated flight loads (40N thrust, 20N weight) to verify structural integrity and ensure an adequate safety factor under normal operating conditions. LS-Dyna simulations assessed landing gear durability under 15m drop tests. Test conditions were iteratively refined to improve energy absorption using foam-encased rods [13]. Drone arms were analysed for natural frequencies and mode shapes to avoid resonance. A motor speed of 1000 rpm (166.67 Hz) was used as the baseline for vibrational safety [14].

CFD analysis was performed using mesh motion to study propeller clearance in hover. Additional flow visualisation over the UAV body was done to understand cruise-phase aerodynamics. No closed-form parametric equations were used. Instead, design parameters were swept on a manually generated grid and each configuration was analyzed in ANSYS under the specified loads and constraints.

### 3.2 Onboard Vision and Subsystem Architecture

Figure 1 shows the avionics stack for cameras, compute boards, RF links and power modules. Perception runs on a Raspberry Pi 5. A Pixhawk 6C handles flight control. They exchange data over a 921 kbit/s serial link using ROS 2 XRCE-DDS. A micro-agent on the Pi publishes the /roi_array detections topic to PX4, where the uXRCE-DDS client converts it into the uORB vehicle_trajectory_waypoint. The bridge by-passes MAVLink parsing for real-time ROI waypoint ingestion. Telemetry topics such as vehicle_gps_position and battery_status stream back to QGroundControl for live path visualisation in Figure 2.

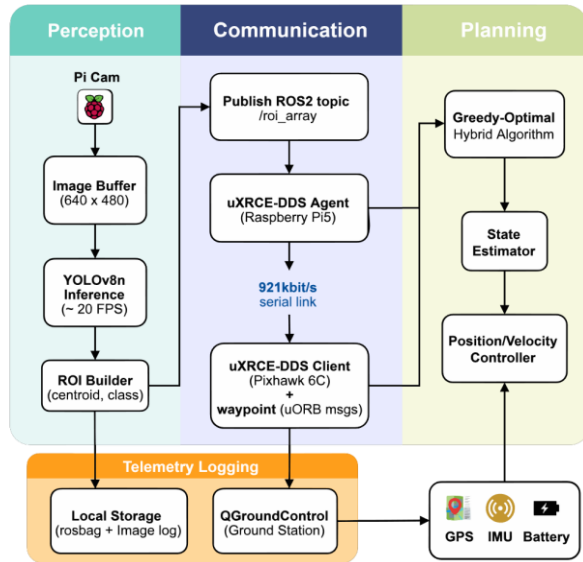

**Fig. 2** Overview of the onboard autonomous navigation pipeline.

On-board inference performance of SSDMobileNetV3-FPN Lite and YOLOv8n was benchmarked using a custom aerial dataset acquired during mission flights. Converting both networks to the NCNN runtime cuts latency by roughly 5× without significant loss in precision; the resulting YOLOv8n in NCNN format runs at about 20 FPS full-resolution and publishes geo-tagged detections to the planner. Fail-safes are embedded in PX4: a GPS bounding-box watchdog triggers Return-to-Home on geo-fence breach, loss of RF link escalates from hover to RTH after 10 s, and battery thresholds at 20 % (warning) and 15 % (auto-land) protect endurance. This vision pipeline, communication bus, and safety logic deliver a fully self-contained SAR-grade autonomous UAV.

### 3.3 Dynamic Path Planning

The proposed dynamic path planning methodology enables the UAV to navigate changing environments efficiently. The algorithm alternates between SWEEP and SERVICE. It first overlays a lawn-mower grid on the geofence; flying this grid alone guarantees full coverage. If no targets are seen, the drone remains in SWEEP and proceeds to the next grid waypoint. Once a ROI appears, the mode switches to SERVICE: the three nearest ROIs are selected by a quick nearest-neighbour test, visited in that order, and removed from the queue. Newly detected ROIs are inserted immediately, preventing stale routes. When the list is empty, the vehicle returns to the grid.

**Algorithm 1** Light-Weight Incremental Greedy (LIG) re-routing algorithm

```
 1  LM ← LawnMowerGrid(P, Δ)              // pre-compute lawn-mower path
 2  S ← ∅;  i ← 0;  mode ← SWEEP
 3  while the mission is active do
 4     S ← S ∪ NewDetections()            // merge novel ROI centroids
 5     mode ← SWEEP if S = ∅ else SERVICE
 6     if mode = SWEEP then
 7        FlyTo(LM[i])
 8        if Reached(LM[i]) then i ← i + 1
 9     else                                // SERVICE mode
10        B ← K nearest points in S to Q   // pure NN scan
11        for each b in B (NN order) do
12           FlyTo(b)
13           S ← S \{b}
14           S ← S ∪ NewDetections()
15           if S = ∅ then break
16        end for
17     end if
18     ClipPathToFence( P )
19  end while
```

## 4  Results and Discussion

The development of an autonomous SAR drone presents a multifaceted engineering challenge. The proposed methodology balances performance, cost, and complexity to achieve a functional prototype. Under the given loading conditions, the structure showed a maximum displacement of 1 mm at the motor arm ends and a peak stress of 0.4 MPa. The computed factor of safety was approximately 3.5, confirming compliance with UAV design requirements.

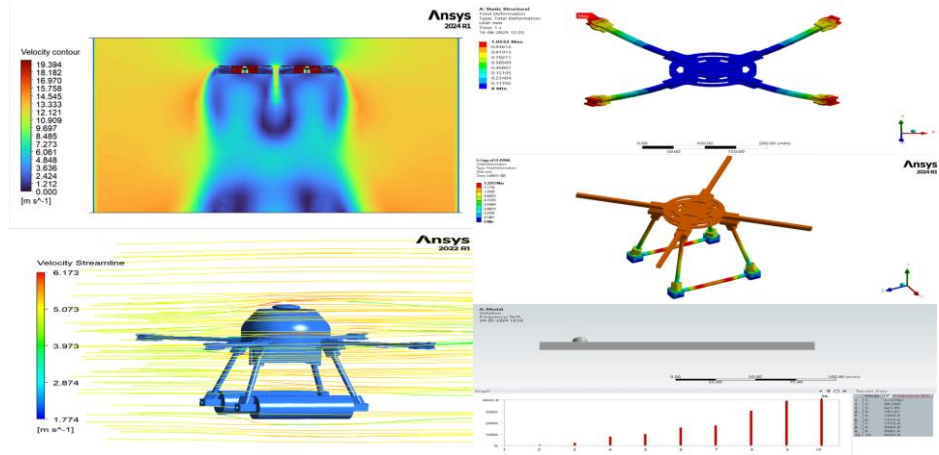

**Fig. 3** Velocity contour (top left) and streamline (bottom left) over UAV; airframe deformation (top right), 15m impact response (mid right), and rotor arm frequency modes (bottom right).

Simulated drop tests confirmed that the landing gear could withstand impacts from 15m. Encasing longitudinal rods with foam significantly enhanced damping and reduced impact stresses, improving resilience on rough terrain, as shown in Figure 3.

**Table 1.** Modes of frequency obtained from harmonic analysis

| Mode | Frequency (Hz) | Deformation axis | Maximum Deformation (in mm) |
|---|---|---|---|
| 1 | 299.67 | lateral axis | 277.67 |
| 2 | 316.31 | vertical axis | 281.07 |
| 3 | 1830.60 | lateral axis | 279.12 |

**Table 2.** CFD results for propeller analysis (mesh motion study)

| Body | Thrust (in N) |
|---|---|
| Propeller 1 | 13.992181 |
| Propeller 2 | 13.970296 |
| Propeller 3 | 13.970296 |
| Propeller 4 | 14.007305 |

The modal analysis confirmed that all the arms' primary vibrational and torsional modes were well above the motor operating frequency of 166.67 Hz, thus eliminating the risk of resonance during flight, as shown in Table 1. Mesh motion study results indicated a cumulative thrust of 56 N (as shown in Table 2), validating a thrust–weight ratio 2. The propeller moment was 0.0083 Nm, confirming torque balance.

Further aerodynamic refinements reduced the drag coefficient from 0.7 to 0.59, improving flight efficiency.

Table 3. Performance comparison of single-shot detectors on Raspberry Pi 5

| Model | mAP | GFLOPs | Size (M) | Latency* |
|---|---|---|---|---|
| SSD-MobileNetV3-FPN Lite | 0.607 | 1.1 | 2.8 | 32 |
| YOLOv8n | 0.734 | 8.7 | 3.2 | 48 |

*Single-thread inference (ms) on Raspberry Pi 5, 640 × 480 input, averaged over 300 frames after NCNN conversion.

Table 4. On-board planning and the time taken by different route-planning algorithms.

| Algorithm | Approach | Time* (in sec) | Mean On-board Latency (in ms) |
|---|---|---|---|
| Nearest Neighbour Heuristic (NNH) | Purely Greedy | 415 | 2 |
| Concorde MILP for TSP | Globally Optimal | 342 | 70 |
| LIG (Proposed) | Hybrid | 355 | 4 |

*Simulated time to visit 25 ROI waypoints in a 600 m × 400 m search box at 5 m s$^{-1}$

Table 3 shows that converting YOLOv8n to NCNN yields a 34% mAP gain over SSD-MobileNetV3-FPN Lite while adding only 16 ms latency. With 8.7 GFLOPs and ~14% more parameters still within Raspberry Pi 5 limits, we use YOLOv8n over SSD-MobileNetV3. As shown in Table 4, the nearest-neighbour heuristic runs in ~2–3 ms but produces the longest tours, whereas Concorde gives the shortest tour yet requires ~70 ms per update on-board [15]. Figure 4 shows rerouting by the proposed algorithm in <5 ms while closing ~85% of the gap to the optimal tour length.

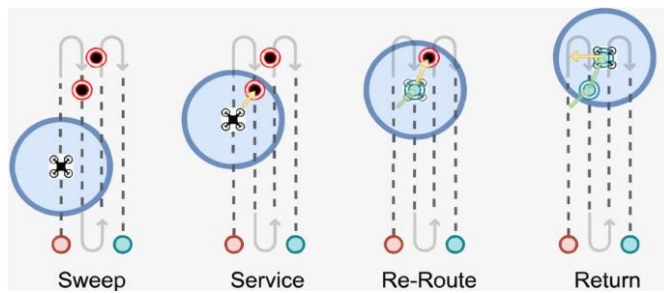

Fig. 4 Re-routing sequence of the proposed planner.

## Conclusion

This work presents a sub-2 kg quadrotor that addresses three bottlenecks: fragile airframe, slow onboard perception, and rigid route planning. Carbon-fibre arms and TPU skids enhance crash tolerance, and FEA and CFD analysis confirm adequate structural stiffness and aerodynamic performance. Converting YOLOv8n to NCNN improves mAP by 34 % over SSD-MobileNetV3-FPN Lite while keeping latency under 50 ms. Detections drive a hybrid greedy–optimal planner that updates tours and reduces detour distance. Future work will improve detection under low visibility, enable coordinated multi-UAV operation, and refine the design for enhanced portability and complex-path navigation.

## References

1. Lyu, M., Zhao, Y., Huang, C., Huang, H.: Unmanned Aerial Vehicles for Search and Rescue: A Survey. Remote Sensing 15(13), 3266 (2023).
2. Ahmad, F., Kumar, P., Khan, Y., Patil, P.: Flow and Structural Analysis of a Quadcopter UAV. Int. J. Adv. Res. Eng. Technol. 11(8), 880–888 (2020).
3. Zhang, P., Zhong, Y., Li, X.: SlimYOLOv3: Narrower, Faster and Better for Real-Time UAV Applications. IEEE Access 8, 92819–92829 (2020).
4. Huang, T., Fan, K., Sun, W., Li, W., Guo, H.: Potential-Field-RRT: A Path-Planning Algorithm for UAVs Based on Potential-Field-Oriented Greedy Strategy to Extend Random Tree. Drones 7(5), Art. 331 (2023).
5. Hassanalian, M., Abdelkefi, A.: Classifications, Applications, and Design Challenges of Drones: A Review. Prog. Aerosp. Sci. 91, 1–21 (2017).
6. Pollet, F., Delbecq, S., Budinger, M., Moschetta, J.-M.: Design Optimisation of Multirotor Drones in Forward Flight. In: Proc. 12th Int. Conf. Unmanned Aerial Systems Engineering, LNCS, vol. 13456, pp. 45–56. Springer, Cham (2021).
7. Benjdira, B., Khursheed, T., Koubaa, A., Ammar, A., Ouni, K.: Car Detection Using Unmanned Aerial Vehicles: Comparison between Faster R-CNN and YOLOv3. In: Proc. 5th Int. Conf. Unmanned Aerial Systems, pp. 112–121. ACM, New York (2018).
8. Cantero, D., Esnaola-Gonzalez, I., Miguel-Alonso, J., Jauregi, E.: Benchmarking Object Detection Deep Learning Models in Embedded Devices. Sensors 22(11), 4205 (2022).
9. Hamza, M.A., Mohsin, M., Khalil, M., Kazmi, S.M.K.A.: MAVLink Protocol: A Survey of Security Threats and Countermeasures. In: Proc. 4th Int. Conf. Digital Futures and Transformative Technologies (ICoDT2), pp. 1–8. IEEE, New York (2024).
10. Nagasawa, R., Mas, E., Moya, L., Koshimura, S.: Model-Based Analysis of Multi-UAV Path Planning for Surveying Post-Disaster Building Damage. Sci. Rep. 11, 18588 (2021).
11. Kyriakakis, N.A., Marinaki, M., Matsatsinis, N., Marinakis, Y.: A Cumulative Unmanned Aerial Vehicle Routing Problem Approach for Humanitarian Coverage Path Planning. Eur. J. Oper. Res. 300(3), 992–1004 (2022).
12. Hayat, S., Yanmaz, E., Bettstetter, C., Brown, T.X.: Multi-Objective Drone Path Planning for Search and Rescue with Quality-of-Service Requirements. Auton. Robots 44, 1183–1198 (2020).
13. Ratuningtyas, A., Azka, M., Riyanti, L., Fatkhulloh, A.: Material Optimisation for Structural Strength of Hexacopter Landing Skid Using Finite Element Method. Int. J. Power Syst. Autom. Technol. 40(1), 234–247 (2023).
14. Öztürk, E.: Unravelling Quadcopter Frame Dynamics: A Study on Vibration Analysis and Harmonic Response. Eur. Mech. Sci. 9(2), 189–195 (2025).
15. Applegate, D.L., Cook, W.J., Rohe, A.: Concorde TSP Solver Benchmarks. Technical Report, University of Waterloo, Canada (1999).